# Few-Shot Generation of Brain Tumors for Secure and Fair Data Sharing

Yongyi Shi, Ge Wang

*Biomedical Imaging Center, Department of Biomedical Engineering, Rensselaer Polytechnic Institute, Troy, NY, USA 12180*

**Leveraging multi-center data for medical analytics presents challenges due to privacy concerns and data heterogeneity. While distributed approaches such as federated learning has gained traction, they remain vulnerable to privacy breaches, particularly in sensitive domains like medical imaging. Generative models, such as diffusion models, enhance privacy by synthesizing realistic data. However, they are prone to memorization, especially when trained on small datasets. This study proposes a decentralized few-shot generative model (DFGM) to synthesize brain tumor images while fully preserving privacy. DFGM harmonizes private tumor data with publicly shareable healthy images from multiple medical centers, constructing a new dataset by blending tumor foregrounds with healthy backgrounds. This approach ensures stringent privacy protection and enables controllable, high-quality synthesis by preserving both the healthy backgrounds and tumor foregrounds. We assess DFGM's effectiveness in brain tumor segmentation using a UNet, achieving Dice score improvements of 3.9% for data augmentation and 4.6% for fairness on a separate dataset.**

Medical imaging is a cornerstone of modern medicine, providing non-invasive visualization of internal features crucial for diagnosis and therapy of many diseases. Recently, deep learning has become increasingly pivotal in computer-aided detection (CAD) systems, enabling earlier and more accurate diagnosis[1]. The effectiveness of deep learning models hinges on access to large, annotated datasets, as these models often contain millions of parameters and continue growing in complexity[2]. In computer vision, significant progress has been driven by open resources such as the ImageNet database[3], which contains over 14 million labeled images[4]. In medical imaging, there are stringent data-sharing restrictions due to ethical considerations, privacy concerns, and regulatory frameworks[5], especially datasets that need to be curated using anonymization techniques[6]. For example, a leading medical image segmentation model has been trained on datasets of around 1.6 million labeled images[7]. However, the availability of certain types of medical images, particularly those associated with rare diseases, remains limited[8]. Moreover, the potential for re-identification attacks poses risks to leakage of patient privacy[9], making hospitals understandably rather cautious about sharing patient data even to collaborators.

Federated learning has emerged as a transformative paradigm, enabling collaborative model training across hospitals without the need for direct data sharing[10-12], thus enhancing privacy protection for individual collaborators' datasets. This decentralized approach has shown significant promise in preserving data confidentiality, especially in sensitive domains such as healthcare[13]. However, recent studies have revealed a vulnerability in federated learning systems: without additional privacy-enhancing mechanisms, these systems can be reverse-engineered, allowing the reconstruction of high-fidelity images from the shared gradient weights of neural networks[14, 15]. Such findings underscore the necessity of further privacy safeguards to patient data. To address these risks, differential privacy (DP) has been proposed as one of the strongest frameworks for mitigating privacy leakage during model training. DP provides formal guarantees by bounding the risk of inferring the individual training samples or reconstructing the original data[16-18]. Despite its potential, the application of DP in federated learning presents a challenging trade-off: while DP ensures privacy protections, it degrades model performance, particularly in highly sensitive applications such as medical imaging[19]. This trade-off between performance and privacy remains a critical barrier to the widespread adoption of DP-enhanced federated learning in healthcare settings.

Synthetic medical images offer a promising avenue for advancing research and clinical practice while protecting patient privacy[20]. Among the prominent methodologies, generative adversarial networks (GANs) used to be popular for medical image synthesis, leveraging the underlying distribution of real images to generate realistic counterparts that can deceive a well-trained discriminator[21]. Although GANs and their variants have been extensively applied to address privacy concerns in medical imaging[22-25], they suffer from major limitations such as failure to capture true image diversity, mode collapse, and unstable training dynamics[26]. In recent studies, diffusion models, which gradually corrupt images into Gaussian noise and reverse this process to generate realistic images[27, 28], have demonstrated marked advantages over GANs, particularly in terms of stability and diversity[29-31]. This has led to a growing interest in diffusion models for medical image synthesis[32-35]. Despite these advancements, diffusion models risk memorizing individual training images and replicating them, a concern exacerbated by typically small sizes of medical datasets[36-37]. Additionally, the heterogenous nature of datasets from multiple institutions, often characterized by demographic and other disparities, raise concerns about fairness[38], which is often overlooked by these models[39].

To enable secure and fair practice of medical image data sharing by addressing demographic biases and label imbalance, this study proposes a decentralized few-shot generative model for synthesizing images by synergizing public and private datasets, taking the synthesis of brain tumor images as an initial example. Current tumor synthesis methods are constrained in terms of tumor deformation[40,41]. To overcome this limitation, we employ a few-shot diffusion model to generate diverse brain tumor images, particularly suited for limited data scenarios, such as rare diseases. Our approach harmoniously integrates generated brain tumors into public normal images, establishing a multi-institute data hub accessible to individual institutions. Using public normal images instead of private patient data, we ensure privacy protection and controllable, high-quality synthesis of images with tumors.

For fairness, a diffusion model is used to generate a synthetic dataset, and models retrained on that dataset are shared, allowing each institute to synthesize brain tumors within their own normal images. This scheme mitigates demographic bias and sample sparsity in model training. Since our approach preserves both the healthy background and tumor foreground, ensuring that the synthesized image quality in these regions remains essentially obeying the underlying distributions of foreground and background features. Hence, retraining the model on synthetic data will maintain accuracy and privacy. Experimental results demonstrate that our method enables data sharing among collaborators while preserving data privacy. Additionally, fairness is also improved as validated on a separate dataset.

**Results**

In this section, a series of experiments are conducted to validate DFGM's performance across various settings. First, its effectiveness in few-shot learning is assessed, particularly in scenarios of rare diseases, such as with limited tumor images. Next, the generated images are used in image segmentation across geographically distributed centers. Finally, fairness is evaluated when sharing synthetic data among different centers.

**Approach and experimental design**

As shown in Figs. 1a and 1b, the few-shot diffusion model is trained to harmonize the public health background with the private tumor foreground. During inference, a similar healthy image is identified by minimizing the L2 distance to the private tumor image, and the two images are then harmonized using the trained diffusion model to synthesize a hybrid image. Fig. 1c illustrates DFGM, designed to address patient privacy, data heterogeneity, and demographic fairness in multi-center datasets. DFGM trains diffusion models at different data centers in a federated learning framework. Then, using task-specific inputs (e.g., segmentation masks) synthetic tumors can be generated and aligned with the data distribution of public health datasets. By generating synthetic tumors exclusively within the public dataset, DFGM eliminates direct access to private datasets. The trained diffusion models then create a large public synthetic database for downstream tasks.

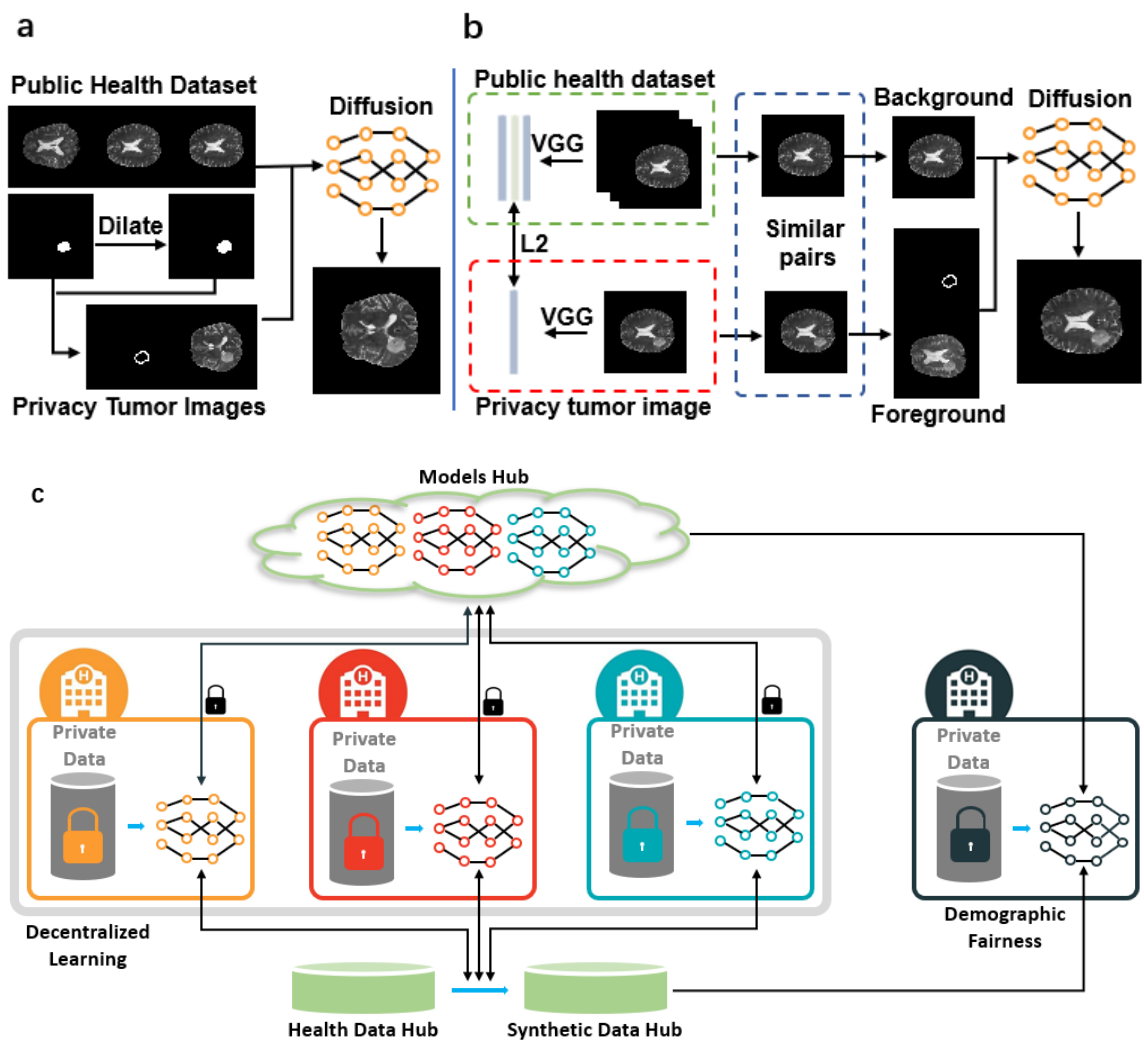

Figure 1. Overview of the method and generated samples. a, Training phase using public health data and private tumor images to develop the diffusion model. b, Testing phase includes extracting a memory bank using a VGG network and finding the closest match by minimizing the L2 distance, followed by harmonizing a public health image background with private tumor foregrounds through the diffusion model to ensure privacy protection. c, decentralized learning application where synthesized privacy-protected images are shared across hospitals, enabling fairness enhancement by generating synthetic tumor images from their own data.

**Few-shot controllable brain tumor image synthesis**

To evaluate generative performance in a limited data scenario, the proposed DFSM is tested in a few-shot learning setting. Fig. 2 presents results from different methods. The tumor image consists of the original tumor, and a corresponding healthy image is identified by selecting the two closest matches based on minimal L2 distances. The DRAEM[42] method directly inserts the tumor into the healthy image, but the tumor boundary appears serrated. AnomalyDiffusion[43] employs a diffusion model to generate tumors, reducing serrated artifacts; however, the tumor shape is not well preserved. In contrast, our method maintains both the tumor and the healthy background, effectively augmenting datasets by altering backgrounds while preserving tumor integrity. Since both the healthy background and tumor foreground are preserved, image quality remains intact. Thus, retraining the diffusion model on this synthetic database ensures complete privacy protection during model sharing, preventing memorization of private data, as all training data remains public.

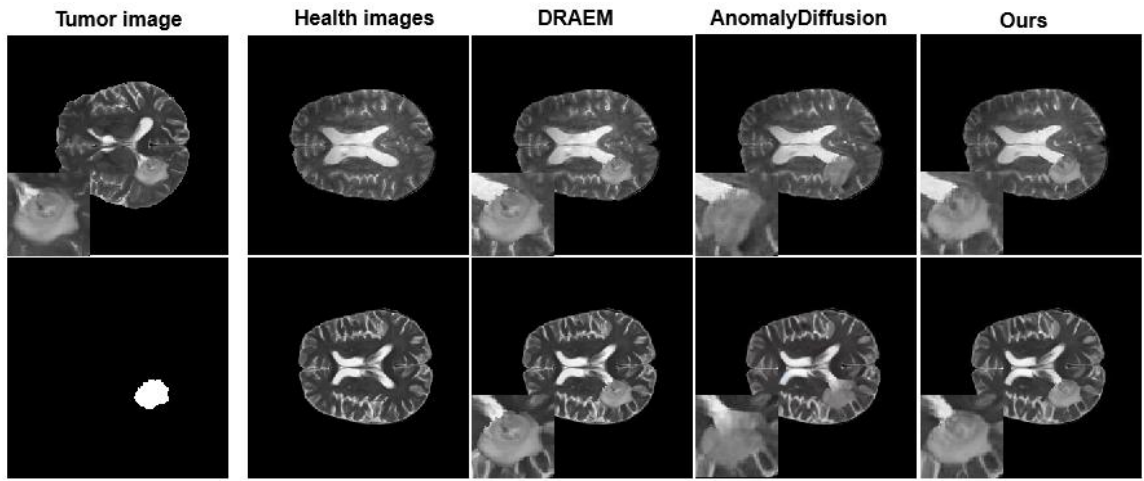

Figure 2. Comparison of our model's synthesized tumor images with existing generation methods, highlighting the superior authenticity and suitability of our results for data augmentation with variable tumor positions and health backgrounds.

**Data augmentation**

In the CBICA dataset, dataset augmentation improves UNet performance, increasing the Dice score from 0.51 to 0.53. The masks in Fig. 3 highlights the one segmentation outcome from the CBICA dataset, demonstrating the alignment between the segmented tumors and the ground-truth. This suggests that the augmented dataset has contributed to refining the model's performance. However, both the results before and after augmentation still show some misalignment with the boundary of the reference mask, likely due to limited training data and the simplicity of the network. Further improvements are needed to improve accuracy.

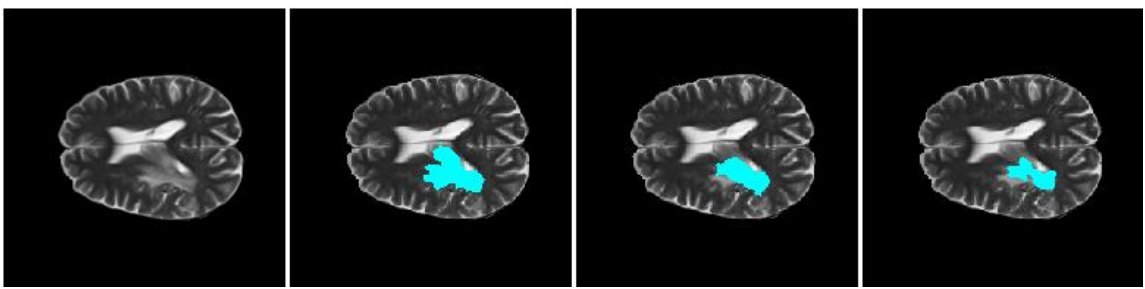

Figure 3. From left to right are the original image, reference mask and the mask predicted by UNet trained on original CBICA dataset and its augmented dataset.

**Fairness evaluation**

We further evaluate the fairness of the augmented dataset using the CBICA dataset and assess the results on the TCIA dataset. To test on the TCIA dataset, the augmented data from CBICA are included in the TCIA training. After applying dataset augmentation, the UNet Dice score improved from 0.65 to 0.68, indicating a notable enhancement in segmentation performance. This suggests that the augmented data helped the model better capture tumor characteristics and improve segmentation accuracy. The masks in Fig. 4 illustrates one segmentation result from the TCIA dataset, demonstrating better alignment between the segmented tumors and the ground-truth masks after augmentation. This suggests that our approach is generalizable and contribute to enhancing the performance of segmentation models across diverse datasets. Further analysis and evaluation are needed to fully understand the long-term impact of these improvements

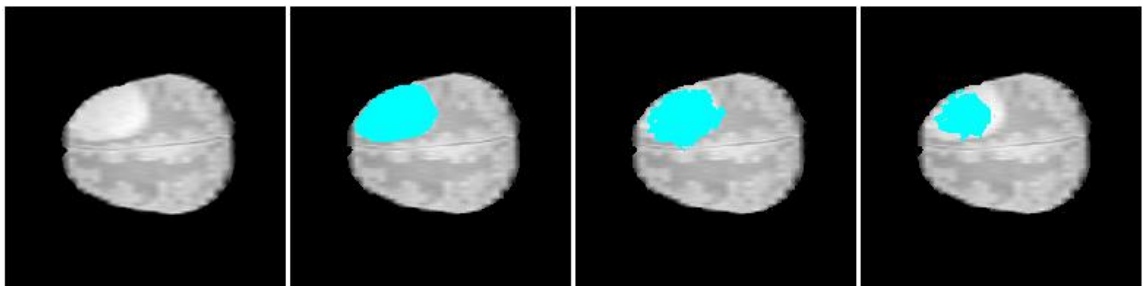

Figure 4. From left to right, the images show the original image, the reference mask, and the masks predicted by the UNet model trained on the original CIA dataset and on the dataset augmented with the CBICIA dataset.

**Discussions**

The DFGM presents a significant advancement in the field of medical image synthesis, particularly for brain tumor images. By combining private tumor data with publicly available healthy images, DFGM enables the creation of synthetic datasets that preserve privacy while improving segmentation performance. The improvements in both data augmentation and fairness demonstrate the potential of this approach to enhance the accuracy and robustness of medical imaging models.

Future work will focus on evaluating the quality of synthetic images through radiologist assessment to ensure clinical viability. Additionally, we plan to assess the performance of more advanced segmentation models, beyond UNet, to determine if the improvements in data augmentation and fairness can be enhanced further. Finally, we will evaluate DFGM's effectiveness on multi-modal data, such as CT images, to test its generalizability across different imaging modalities. These efforts will help improve the model's robustness and its applicability in real-world clinical settings.

**Methods**

**Data collection**

We collect glioblastoma (GBM) data from the Brain Tumor Segmentation Challenge 2020 (BraTS20) training dataset, with MRI scans acquired using various clinical protocols and scanners from different institutions, specifically from the Center for Biomedical Image Computing and Analytics (CBICA) and the Cancer Imaging Archive (TCIA), totaling 309 scans. Each scan includes four MRI modalities: native (T1), post-contrast T1-weighted (T1Gd), T2-weighted (T2), and T2 Fluid-Attenuated Inversion Recovery (T2-FLAIR), with ground truth annotations identifying three tumor sub-regions, including tumor core, enhancing tumor, and edema. Evaluation focuses on whole tumor segmentation in the T2 modality. The dataset is split at the patient level into training and test sets, with CBICA contributing 103 scans for training and 26 for testing, while TCIA provides 134 scans for training and 33 for testing. Since analysis is conducted in two dimensions, 2D tumor-containing slices are extracted at five-slice intervals to reduce redundancy, resulting in 483 slices for training and 129 for testing from CBICA, and 562 slices for training and 141 for testing from TCIA. In addition, 825 healthy slices without tumors are collected as a public health dataset.

We use tumor-bearing training data from CBICA along with public health data to train our diffusion model, generating tumor-bearing data by identifying similar pairs from the health dataset. The diffusion model is then applied to augment the dataset, and the augmented data is incorporated into both the CBICA and TCIA test sets to evaluate the effects of data augmentation and fairness.

**Network architecture**

As shown in Fig. 1a, both the public health dataset and private tumor images are used to train our diffusion model. For tumor images, the tumor mask is dilated to define the tumor boundary, allowing the diffusion model to generate tumor boundaries that seamlessly blend with the surrounding normal tissue. During inference, for each private tumor image, a similar counterpart is identified from the public health dataset. VGG is used to extract feature vectors, and the L2 distance is calculated to determine the closest match, selecting the most similar image based on the minimal L2 distance. The tumor is then directly inserted into the selected healthy image, serving as the input to the diffusion model. The tumor boundary is also provided as a mask input. The model generates an output image where the tumor foreground is harmonized with the healthy background, ensuring a smooth transition between the two regions.

**Few-shot diffusion model**

The model focuses on learning the differences within the boundary mask regions rather than modeling the entire dataset, which makes it well-suited for few-shot learning. We

1117

adopt the framework from AnomalyDiffusion to generate a large amount of tumor data aligned with tumor masks, learning from only a few tumor samples.

The inputs to our model include an image $\boldsymbol{y}$, where the tumor foreground from a private dataset is inserted into a healthy background, along with a tumor boundary mask $\boldsymbol{m}$. The output is a harmonized image with the tumor, where both the background and foreground are consistent with the input image $\boldsymbol{y}$. Only the content within the boundary mask region is generated.

To provide tumor location information, spatial embedding $\boldsymbol{e}$ is encoded from the boundary mask $\boldsymbol{m}$. Given this spatial embedding as a condition, and an image $\boldsymbol{y}$, we generate the tumor image through the blended diffusion process:

$$\boldsymbol{x}_{t-1} = p_\theta(\boldsymbol{x}_{t-1}|\boldsymbol{x}_t, \boldsymbol{e}) \odot \boldsymbol{m} + q(\boldsymbol{y}_{t-1}|\boldsymbol{y}_0) \odot (1 - \boldsymbol{m}) \qquad (1)$$

**Learning of downstream tasks**

We focus on tumor segmentation as the downstream task. After training a well-learned image generator, synthetic medical images can be generated using either public or self-collected healthy images, along with the corresponding tumor and mask data. In our experiments, to ensure a fair comparison between synthetic and real samples, we used the same UNet model for segmentation across different 2D image sets. During training, we used the Adam optimizer with a learning rate of 0.001 and employed a combination of Binary Cross-Entropy with Logits Loss (BCEWithLogitsLoss) as the loss function for the segmentation task.

**Quantitative metrics**

The Dice score is used to evaluate segmentation performance. It measures the overlap between the ground-truth mask and the segmented results and is defined as:

$$Dice = \frac{2 \cdot |A \cap B|}{|A| + |B|} \tag{2}$$

where A represents the ground-truth mask, and B represents the segmented result. The score ranges from 0 (no overlap) to 1 (perfect overlap).

Page 14 of 17